\documentclass{article}
\usepackage{ijcai17}
\usepackage{graphicx}

\usepackage{times}
\usepackage[linesnumbered,lined,commentsnumbered, ruled,vlined]{algorithm2e}
\usepackage{bm}
\usepackage{amsmath}
\usepackage{amssymb}
\usepackage{amsthm}
\usepackage[font=small]{caption}
\DeclareMathOperator{\E}{\mathbb{E}}

\DeclareMathOperator{\M}{\mathcal{M}}
\DeclareMathOperator{\D}{\mathcal{D}}
\DeclareMathOperator{\SL}{\mathcal{S}}
\DeclareMathOperator{\A}{\mathcal{A}}
\DeclareMathOperator{\B}{\mathcal{B}}
\DeclareMathOperator{\I}{\mathcal{I}}
\DeclareMathOperator{\SEN}{\mathcal{SEN}}

\newcommand\Mark[1]{\textsuperscript#1}

\newtheorem{theorem}{Theorem}[section]

\usepackage{graphicx}

\title{Multimodal Storytelling via Generative Adversarial Imitation Learning}
\Large{
\author{Zhiqian Chen\Mark{1}, Xuchao Zhang\Mark{1}, Arnold P. Boedihardjo\Mark{2}, Jing Dai\Mark{3}, Chang-Tien Lu\Mark{1} \\ 
\Mark{1}Computer Science Department, Virginia Tech, Falls Church, Virginia  \\
\Mark{2}U. S. Army Corps of Engineers\\
\Mark{3}Google Inc.\\
\Mark{1}\{czq,xuczhang,ctlu\}@vt.edu,\Mark{2}arnold.p.boedihardjo@erdc.dren.mil, \Mark{3}jddai@google.com
}
}

\begin{document}

\maketitle

\begin{abstract}
Deriving event storylines is an effective summarization method to succinctly organize extensive information, which can significantly alleviate the pain of information overload. The critical challenge is the lack of widely recognized definition of storyline metric. Prior studies have developed various approaches based on different assumptions about users' interests. These works can extract interesting patterns, but their assumptions do not guarantee that the derived patterns will match users' preference. On the other hand, their exclusiveness of single modality source misses cross-modality information. This paper proposes a method, multimodal imitation learning via generative adversarial networks(MIL-GAN), to directly model users' interests as reflected by various data. In particular, the proposed model addresses the critical challenge by imitating users' demonstrated storylines. Our proposed model is designed to learn the reward patterns given user-provided storylines and then applies the learned policy to unseen data. The proposed approach is demonstrated to be capable of acquiring the user's implicit intent and outperforming competing methods by a substantial margin with a user study.
\end{abstract}

\section{Introduction}

\begin{figure}[!hpbt]
    \centering
    \includegraphics[width=3.2in]{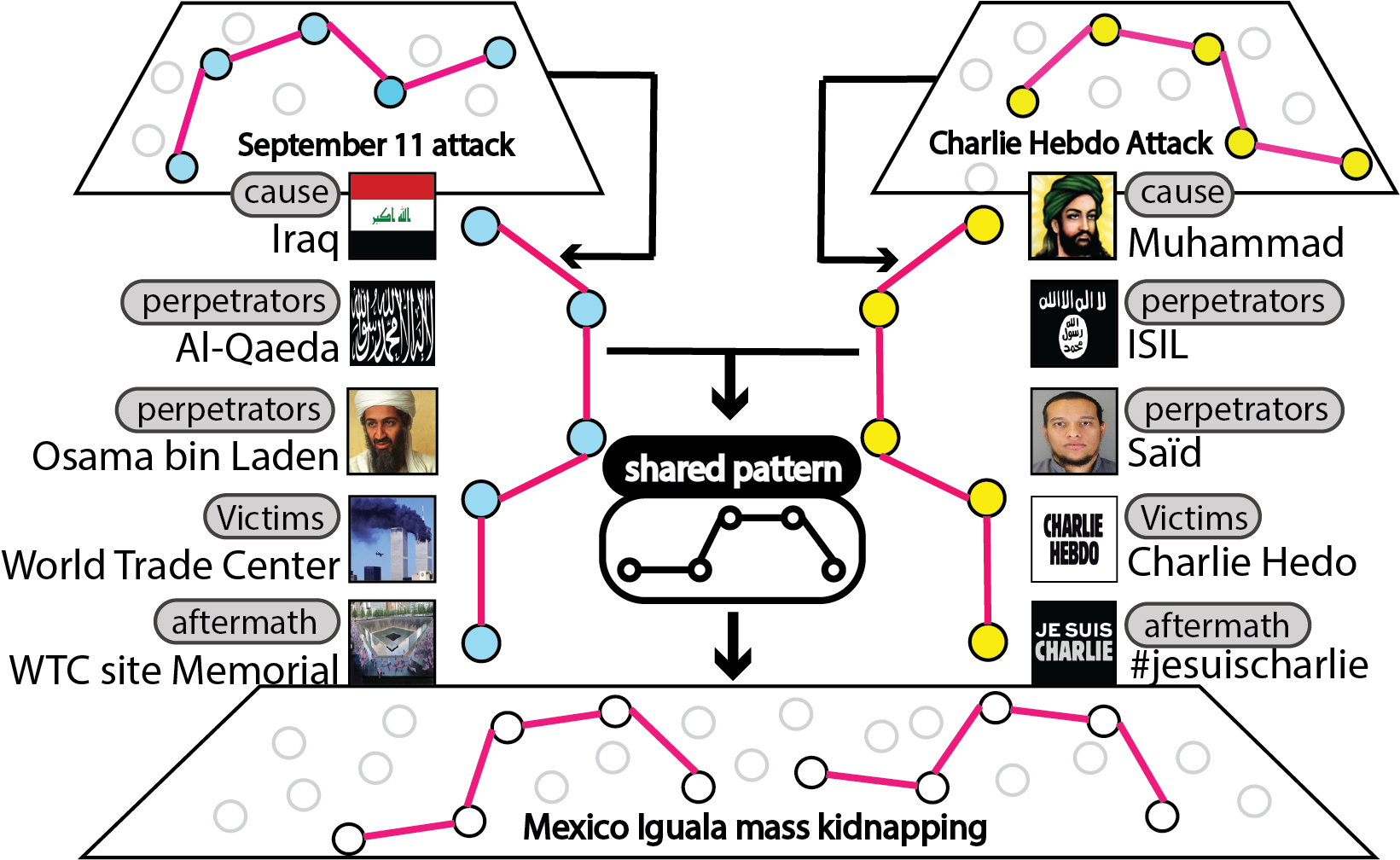}
    \caption{Imitation Storytelling}
    \label{MIS}
\end{figure}
As the Internet becomes more pervasive, information overload becomes increasingly more severe. Even with the help of search engines such as Google and Yahoo, people cannot easily understand a series of coherent news events. For example, a person who desires to learn about the \textit{2016 Presidential Election} needs iteratively search through several keywords many times and review numerous news documents so that he or she can generate a cohesive picture, e.g., knowledge graph. \textbf{Storytelling} is an efficient way to solve this issue of information overload. By inferring the entity nodes connections, the original documents can be represented through a knowledge graph which consists of a set of storylines. Current works  \cite{kumar2008algorithms,hossain2012storytelling,shahaf2012metro,shahaf2012trains,lin2012generating} mainly focus on this task but employ strict assumptions that may not capture meaningful stories. For instance, \cite{shahaf2010connecting} argues that maximizing the weakest link makes good storylines, which can be reduced to density-based clustering. However, to model users' preferred stories, it requires to understand the evolution patterns, not merely to keep strong coherence.

Specifically, existing research in this area suffers from several shortcomings: \textbf{(1) Strong assumptions can lead to poor storylines.} Most related works manually design coherence metrics which are directly assumed to be associated with good storylines. However, a high consistency does not guarantee story quality, since good story might have other properties such interest, novelty, or user-preferred style. Therefore, the coherence based metric is not sufficient for modeling storylines. \textbf{(2) Lack of multimodal learning}. Existing works only focus on unimodal data such as Text Storytelling or Visual Storytelling, overlooking the cross-modality information. Multimodal learning can find entity linkages that a unimodal may miss. \textbf{(3) Absence of benchmark dataset}: Few prior works are reported to provide a publicly available dataset for imitation storytelling.

This paper focuses on directly imitating user-provided storylines rather than designing any indirect measures. The basic idea is to learn the connectivity features and structure in storylines so that the agent can reveal similar stories in other domains. This approach is illustrated in Figure \ref{MIS}. The \textit{September 11 attack} event contains a storyline which shows the key entities related to the cause, perpetrators, victims, and the aftermath. Similarly, the event \textit{Charlie Hebdo Attack} also has a story of a similar type. We argue that the two similar storylines share the same structure in a certain embedding space. Therefore, we can reveal similar stories in another event domain(\textit{Mexico Iguala mass kidnapping} at the bottom). The inherent benefit of utilizing multimodal data is that humans often make inferences between the images and texts so as to resolve ambiguities.

Deep reinforcement learning can learn multi-step decisions. However, a critical difficulty in reinforcement learning is designing a reward function for optimizing the agent. Unlike game application in which there exists a responsive environment, it is dramatically difficult to design a reward function for storytelling due to the unavailability of such responses. Instead of proposing reward function, we introduce a typical Inverse Reinforcement Learning(IRL), imitation learning, to learn the latent policy. Imitating from demonstrations is a strategy in which an agent learns a hidden policy for a dynamic environment by observing demonstrations delivered by an expert agent. While IRL is often with two issues: instability and implicit policy. To solve these problems, Generative Adversarial Networks(GAN)\cite{goodfellow2014generative} mechanism is employed to solve the instability issue. GAN yields an internal generator model which can output policy explicitly after training, which addresses the second issue. Therefore, it is promising to integrate IRL with GAN to learn policy from users' demonstrations.

Different from previous work, our study treats storytelling as an imitation learning. Specifically, a policy is acquired from one event domain, and then transfer the policy to learn a storyline from another event. Furthermore, our work takes full advantage of multimodal data to improve the imitation performance. In this paper, we define a Multimodal Imitation Storytelling Task(MIST), and then propose a multimodal generative adversarial method that derives latent policy behind users' demonstrations without explicitly designing a reward function. The main contributions as follows:
\begin{itemize}
    \item \textbf{Proposing an imitation learning method for storytelling}: To avoid the difficulty in designing reward function for storytelling, we enforce generative adversarial model on imitation learning. Using this learning strategy, the model can robustly model latent connectivity patterns.
    \item \textbf{Designing a multimodal model integrated with GAN based imitation learning}:  Inspired by human's ability to link multiple entities through visual similarity, we propose a multimodal method across textual and visual modality with imitation learning. Our model learns reward functions from these two modalities and their correlation.
    \item \textbf{Creating a benchmark dataset for multimodal imitation storytelling}: A new multimodal storytelling dataset is collected from multiple attacks and civil unrest events. Under several selected topics, storylines are manually extracted and validated. Both texts and images are included in our dataset.
\end{itemize}

The rest of this paper is organized as follows. Section \ref{rw} reviews the related works. A detailed description of the proposed method is given in Section \ref{method}. Experiments on multiple public datasets and case study are presented in Section \ref{exp}. We conclude the paper in Section \ref{conclusion}.

\section{Related Works}\label{rw}
\textbf{Storytelling:} The storyline generation problem was first studied by Kumar et al. \cite{kumar2008algorithms} as a generic redescription mining technique, by which a series of re-description between the given disjoint and dissimilar object sets and corresponding subsets are discovered. Storytelling is an efficient way to solve the issue of information overload. By extracting critical and connected entities, the original document is structurally summarized. Current works contain two categories: Textual Storytelling\cite{kumar2008algorithms,hossain2012storytelling,fang2011rex,voskarides2015learning,lee2012joint,shahaf2012metro,shahaf2012trains,lin2012generating} and Visual Storytelling\cite{kim2014joint,park2015expressing,wang2012generating}. Few works are reported to extract storylines based on both text and image. Current methods often suggest assumptions between good storyline and explicit metrics, such as average similarity or weakest similarity of all the neighbor nodes. However, these assumptions limit the generating meaningful stories since a user may have unique notions of good storylines. A few researchers employ Latent Dirichlet Allocation(LDA)\cite{zhou2015unsupervised_a,huang2013optimized} to extract stories in unsupervised fashion. However, it is difficult for LDA to accurately model sequential data. 

\textbf{Reinforcement Learning:} Starting from AlphaGo \cite{silver2016mastering} and Atari\cite{mnih2015human}, numerous game applications \cite{lample2016playing,he2016deep_a,oh2015action,narasimhan2016improving} enjoy the property of deep reinforcement learning in imitating sequential patterns. However, it is often difficult to design rewards function, especially for this real world problem. Other possible solutions include behavior cloning\cite{pomerleau1991efficient}, and Inverse Reinforcement Learning(IRL)\cite{russell1998learning,ng2000algorithms} which can derive the underlying cost function under which the expert data is uniquely optimal. Unfortunately, behavior cloning learns a policy as a supervised learning problem over state-action pairs from expert trajectories and required large amounts of data because of compounding error caused by covariate shift. While IRL has stability issue and does not explicitly tell us how to act. Several works\cite{yu2017seqgan,ho2016generative} employ Generative Adversarial Networks(GAN) solve IRL issue. Hence we leverage GAN and IRL to effectively imitate user-provided storylines.

\textbf{Multimodal Learning} is to derive the patterns in associated cross-modality information. An example is that readers often estimate the relationship among news articles using the top images inside because if the images contain the same entity, they are probably involved in the same event. However, current works\cite{kiros2014unifying}\cite{srivastava2012multimodal}\cite{ngiam2011multimodal} focus on joint representation learning, paying little attention to sequence problem.

Different from the previous study, our work treats storytelling as a combination problem of imitation learning and multimodal learning. By employing GAN based imitation learning, our proposed model can learn and show the hidden policy. Moreover, this work takes full advantage of joint constraint on cross-modality data to improve the imitation performance.

\section{Multimodal Imitation Storytelling}\label{method}
This section formally defines the task of imitation storytelling and then describes our proposed MIL-GAN model. In particular, a multimodal and GAN based imitation learning is elaborated in \ref{MIL-GAN}, the multimodal method we applied is introduced in \ref{ML}.

\subsection{Problem Setup}
Based on a set of event document-storyline pairs \{$\D$, $\SL$\}, our goal is to reveal the user generated policy $\pi^{G_{user}}$. Each $\D$ is a documents collection and each $\SL$ consist of several entity nodes \{$e$\}

Let $\M=\{txt, img, mm \}$ be the types of modalities appearing in $\D$ and $\SL$, which denote textual data, image and multimodal relationship between respectively. In the proposed method, a generator $\pi^{G_{\bm \theta}}$ controlled by multimodal parameters $\bm \theta=\{\theta^{txt},\theta^{img},\theta^{mm}\} $ is used to approximate the users' policy generator $\pi^{G_{user}}$. 

Following REINFORCE algorithm\cite{williams1992simple}, the collected rewards $R_{T}$ along the sequence starting from the initial state $\bm e_{0} =\{e^{txt},e^{img},e^{mm}\}$ are maximized so as to optimize the generator. Utilizing GAN model, a discriminator $\D_{\phi}$ yields rewards as the state score of the proposed generator. Since the $Q$ value function is influenced by both the generator and discriminator, let $Q^{G_{\theta} }_{D_{\phi}}$ denote the $Q$ function. The learning process iteratively updates the objective function via gradient w.r.t. the parameters $\bm \theta$ until convergence.

\subsection{Imitation Learning via GAN}\label{MIL-GAN}
The intermediate rewards are set to zero for storytelling task because it's difficult to evaluate the generated storyline until it's complete. Following \cite{sutton1999policy},the objective of policy $\pi \left(s_{t}|\SL_{1:t-1}\right)$ is to generate a sequence starting from $\bm s_{0}$ to maximize the expected reward:
\begin{equation}
J\left( \theta \right) =
\E\left[\bm{R_{T}}\vline \bm{e_{0}},  \bm{\theta}  \right]
\end{equation}

Because multimodal constraints involves several separate relationship embedded in text, image and their bi-directional coherence, the objective function is formulated as:

{\small
\begin{multline}\label{loss}
J\left(\bm \theta \right)=\E\left[\sum_{i\in \M}\lambda^{i} R_{T}^{i} \vline \bm{e_{0}},  \bm{\theta}  \right] =\sum_{i \in \M}\lambda^{i}\E^{i}\left[{R_{T}^{i}}\vline {e_{0}^{i}},  {\theta^{i}}  \right]\\
=\sum_{i \in \M}\lambda^{i}\sum _{s_{1}\in \SL^{i}} \pi^{G_{\theta^{i}}}\left( s_{1}\vline e_{0}^{txt}\right) \cdot Q^{G_{\theta^{i}} }_{D_{\phi^{i}} }\left( e_{0}^{i},s_{1}\right)
\end{multline}}%

\begin{figure}[!hpbt]
    \centering
    \includegraphics[width=3in]{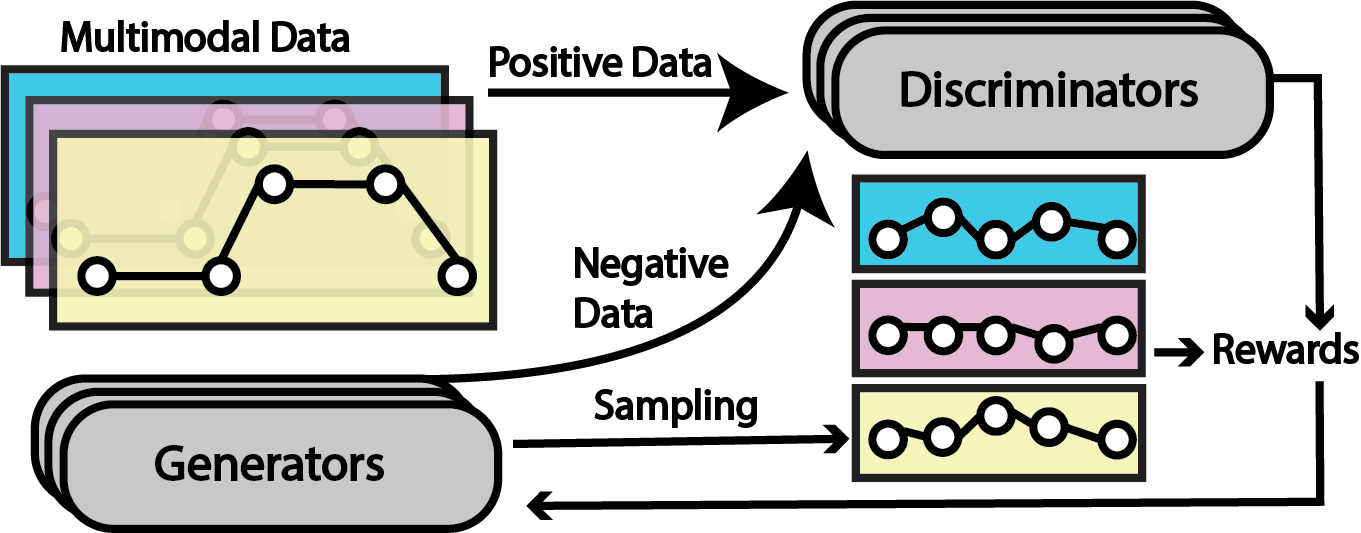}
    \caption{Illustration of Multimodal GAN: Large arrows show the data flow in GAN training, while small arrows indicate the process inside policy gradient}
    \label{mmm}
\end{figure}

where $Q^{G_{\theta} }_{D_{\phi}}\left(state, action\right)$ is the action-value function that estimates the expected accumulative reward from initial state $e_{0}$. The $Q$ value is from the rating on sequence generated by $\pi^{G_{\bm \theta}}$. When the discriminator consider the generated sequences real, the $Q$ value will increase. The basic idea of estimating $Q$ value is to treat it as the probability of the sequence being real considered by the discriminator, i.e., $Q^{G_{\theta} }_{D_{\phi} } =D_{\phi } $. Due to the model limitation, only the sequence of fixed length can be evaluated by discriminator. However, incomplete sequence does provide partial information for rating. To estimate this incomplete portion of the sequence, the action-value function value is assigned with the expected rewards obtained from LSTM sampling controlled by $\pi^{G_{\theta}}$. Therefore, the $Q^{G_{\theta} }_{D_{\phi} }\left( \SL_{1:t-1}, s_{t}\right)$ value is estimated as:
{\small
\begin{multline}\label{mc}
\begin{cases}
D_{\phi }\left( \SL\right) & \text{if $len(\SL)=T$} \\
\E_{\pi^{G_{\theta}}}\big[ D_{\phi }\left( LSTM(\SL, \pi^{G_{\theta}})\right)\big]  & \text{$else$}\\
\end{cases}
\end{multline}}

By using such $Q$ values, the generator can update the objective function w.r.t. its parameters. Once generator finishes update, the sampling sequences mixed with the real data are fed into discriminator. Following the typical rules of GAN and an improvement\cite{wgan} that the original log operation of discriminator is removed, the discriminator will then improve the function:
\begin{equation}\label{gan}
\resizebox{.9\hsize}{!}{$\min _{\phi }-\E_{Y\sim P_{data}}\left[ D_{\phi }\left( \SL\right) \right] -\E_{Y\sim \pi^{G_{\theta }}}\left[ \left( 1-D_{\phi }\left( \SL\right) \right) \right]$}
\end{equation}

As illustrated in Figure \ref{mmm}, the GAN iteratively updates the multimodal generators using Formula \ref{mc}, and improves the multimodal discriminators by Formula \ref{gan}. Following policy gradient theorem(episodic case) \cite{sutton1999policy}, the gradient of the objective function $J(\theta)$ w.r.t. the  parameters $\theta$ can be expressed as:
\begin{theorem}[Multimodal Policy Gradient Theorem]
    \label{theorem}
    The gradient of the objective function $\nabla _{\theta }J \left(\bm \theta\right)$ w.r.t. the parameter $\bm \theta$ via policy gradient is:
    \[ \sum_{i \in \M} \lambda^{i}\cdot\E_{G_{\theta_{i}}}\left[ \sum_{s_{t}\in \SL_{i}}\nabla _{\theta^{i} }\pi^{G_{\theta_{i} }}\left( s_{t}\vline \SL_{1:t-1}\right)\cdot Q^{G_{\theta^{i}} }_{D_{\phi^{i}} }\left( \SL_{1:t-1},s_{t}\right) \right]  \]
\end{theorem}
\begin{proof}[Proof of Theorem \ref{theorem}]
    Let $v$ denote the state value function. To keep the notation simple, we leave it implicit in all cases that $v$, $\pi$ and $Q$ are controlled by the generator $G^{\theta_{i}}$. For each single modality, due to that intermediate rewards are set to zero, and state transferring probability $\gamma$ is one-hot deterministic, we have:
    \begin{equation}\label{qfunc}
    Q\left( \SL_{1:t-1},s_{t}\right) =0+\sum _{s'\in \SL}\gamma _{ss'}v\left(s'\right) =v\left( \SL_{1:t}\right)
    \end{equation}
    \begin{equation}\label{ufunc}
    v\left( \SL_{1:t-1}\right) =\sum _{s_{i}\in \SL}\pi\left( s_{t}\vline \SL_{1:t-1}\right) Q\left( \SL_{1:t-1},s_{t}\right)
    \end{equation}
    Sum over modalities $\M$ on Eq. (\ref{ufunc}) is exactly Eq. (\ref{loss}), i.e., $    \nabla _{\theta }J=\sum_{i \in \M} \lambda^{i}\cdot\nabla _{\theta^{i} }v\left( s_{0}\right) $. Substitute $v$ in Eq. (\ref{ufunc}) and $Q$ in (\ref{qfunc}) iteratively:
    {\small  
    \begin{align*}    
    \nabla _{\theta }&J=\sum_{i \in \M} \lambda^{i}\cdot\nabla _{\theta^{i} }v\left( s_{0}\right)&\\
    =&\sum_{i \in \M} \lambda^{i}\cdot\nabla _{\theta ^{i}}\left[\sum _{s_{1}\in \SL}\pi \left( s_{1}\vline s_{0}\right)Q\left( s_{0},s_{1}\right) \right] &\\
    =&\sum _{i\in\M}\lambda ^{i}\sum _{s_{1}\in \SL}\left[ \nabla _{\theta }\pi \left( s_{1}\vline s_{0}\right) Q\left( s_{1}\vline s_{0}\right) +\pi \left( s_{1}\vline s_{0}\right) \nabla _{\theta }Q\left( s_{0},s_{1}\right)\right]&\\
    =&\sum _{i\in\M}\lambda ^{i}\sum _{s_{1}\in \SL}\left[ \nabla _{\theta }\pi \left( s_{1}\vline s_{0}\right) Q\left( s_{1}\vline s_{0}\right) +\pi \left( s_{1}\vline s_{0}\right) \nabla _{\theta }v\left( \SL_{1:1}\right)\right]&\\
    =&\sum _{i\in\M}\lambda ^{i}\sum _{s_{1}\in \SL}\Big[ \nabla _{\theta }\pi \left( s_{1}\vline s_{0}\right) Q\left( s_{1}\vline s_{0}\right) +&\\
    &\pi \left( s_{1}\vline s_{0}\right)\nabla _{\theta ^{i}}\big[\sum _{s_{1}\in \SL}\pi\left( s_{1}\vline s_{0}\right) Q\left( s_{0},s_{1}\right) \big] \Big] &\\
    =&\sum _{i\in \M}\lambda ^{i}\sum _{s_{1}\in \SL}\Big[ \nabla _{\theta }\pi \left( s_{1}\vline s_{0}\right) Q\left( s_{1}\vline s_{0}\right) +&\\ 
    &\pi \left( s_{1}\vline s_{0}\right)\sum _{s_{1}\in \SL}\left[ \nabla _{\theta }\pi \left( s_{1}\vline s_{0}\right) Q\left( s_{1}\vline s_{0}\right) +\pi \left( s_{1}\vline s_{0}\right) \nabla _{\theta }Q\left( s_{0},s_{1}\right) \right] \Big]
    \end{align*}}%
    We can re-write accumulative policy output as probability: 
    \begin{equation}\label{Pr}
    \resizebox{.9\hsize}{!}{$
        Pr(\SL_{1:t} | s_{0})=\prod_{i}^{t}\pi(s_{i}|\SL_{1:i-1})=\pi \left( s_{1}\vline s_{0}\right)\pi \left( s_{2}\vline \SL_{1:1}\right)..\pi \left( s_{t}\vline \SL_{1:t-1}\right)
    $}
    \end{equation}
    After repeating unrolling and apply Eq. (\ref{Pr}), it is then immediate that:

    {\tiny  
    \begin{align*}\label{mergeprob}
    \nabla _{\theta }J=&\sum _{i\in\M}\lambda ^{i}\sum_{s_{1}\in \SL}\Big[ \nabla _{\theta }\pi \left( s_{1}\vline s_{0}\right) Q\left( s_{1}\vline s_{0}\right) &\\
    +&Pr\left( \SL_{1:1}\vline s_{0}\right)\sum _{s_{1}\in \SL}\nabla _{\theta} \pi\left( s_{1}\vline s_{0}\right) Q\left( s_{1}\vline s_{0}\right) &\\
    +&\pi \left( \SL_{1:2}\vline s_{0}\right) \nabla _{\theta}Q\left( s_{0},s_{1}\right)  \Big]&\\
    =\sum _{i\in \M}&\lambda^{i}\Big[ \sum ^{T}_{t=1}\sum _{\SL_{1:t-1}}Pr\left( \SL_{1:t-1}\vline s_{0};\pi \right) \sum _{s_{t}\in \SL}\nabla _{\theta }\pi \left(s_{t}\vline \SL_{1:t-1}\right) Q\left( \SL_{1:t-1},s_{t}\right)&\\ 
    +&\sum _{s_{t}\in \SL}\pi\left( s_{t}\vline \SL_{1:t-1}\right) \nabla _{\theta }Q\left( \SL_{1:t-1},s_{t}\right)  \Big]
    \end{align*}}%
    In storytelling, each node can only be assigned with one entity, the policy function in the last term given a specific $s_{t}$ reach $\lim\limits_{T\to\infty} \pi\left( s_{t}\vline \SL_{1:t-1}\right)=\inf \pi\left( s_{t}\vline \SL_{1:t-1}\right) = 0$. Accordingly, the above equation becomes:
    {\tiny
    \begin{align*}
    &\nabla _{\theta }J=\sum _{i\in \M}\lambda^{i}\sum ^{T}_{t=1}\sum_{\SL_{1:t-1}}Pr\left( \SL_{1:t-1}\vline s_{0};\pi \right) \sum _{s\in \SL}\nabla _{\theta }\pi \left( s_{t}\vline \SL_{1:t-1}\right) Q\left( \SL_{1:t-1},s\right)&\\
    &=\sum_{i \in \M} \lambda^{i}\cdot\E_{G_{\theta_{i}}}\left[ \sum_{s_{t}\in \SL_{i}}\nabla _{\theta^{i} }\pi^{G_{\theta_{i} }}\left( s_{t}\vline \SL_{1:t-1}\right)\cdot Q^{G_{\theta^{i}} }_{D_{\phi^{i}} }\left( \SL_{1:t-1},s_{t}\right) \right] 
    \end{align*}}%
\end{proof}

Using weighted likelihood \cite{sutton1999policy}, the expectation inside Formula \ref{theorem} can be decomposed using unbiased estimation. To keep the notation simple, we leave it implicit in all cases that $\pi$ is a function of $s_{t}$ given $\SL_{1:t-1}$, and $Q^{G_{\theta} }_{D_{\phi} }$ is a function of $s_{t}$ and $\SL_{1:t-1}$. Thus, the gradient can be estimated as:
{\small  
\begin{align*}
\nabla _{\theta }J \left(\bm \theta\right)&\simeq \sum_{i \in \M} \lambda^{i}\cdot \dfrac {1}{T}\sum ^{T}_{t=1}\sum _{s_{t}\in \SL}\nabla _{\theta^{i} } G_{\theta^{i} } \cdot Q^{G_{\theta^{i}} }_{D_{\phi^{i}} }&\\
&= \sum_{i \in \M} \dfrac {\lambda^{i}}{T}\sum ^{T}_{t=1}\sum _{s_{t}\in \SL}G_{\theta^{i} } \nabla _{\theta^{i} }\log G_{\theta^{i} }\cdot Q^{G_{\theta^{i}} }_{D_{\phi^{i}} } &\\
&= \sum_{i \in \M} \dfrac {\lambda^{i}}{T}\sum ^{T}_{t=1} \E_{s_{t}\sim G_{\theta^{i} }} \left[  \nabla _{\theta^{i} }\log G_{\theta^{i} }\cdot Q^{G_{\theta^{i}} }_{D_{\phi^{i}} } \right]
\end{align*}}%

The gradient can then be applied on the previous parameters as: $\theta^{i} \leftarrow \theta^{i}+\alpha\nabla _{\theta^{i} }J\left( \theta^{i} \right)$, where $\alpha$ is learning rate. Algorithm \ref{mis} presents full details of the proposed method. First, we pre-train $G^{i}$ on input set $\SL$. Then, the generator and discriminator are trained alternatively and periodically. When training the discriminator, positive examples are from the given dataset, while negative examples are sampled from the generator.
\begin{algorithm}[!h]
    \caption{Multimodal Imitation Storytelling}
    \label{mis}
    \SetAlgoLined
    \KwIn{initialize generator policy $\pi^{G_{\theta}}$ and discriminator $D_{\phi}$ with random weights; multimodal storyline dataset $Data=\{E(ntity)_{1:T}, I(mg)_{1:T}\}$; event documents as knowledge base $\D$}
    \KwOut{Generator $\pi^{G_{\theta}}$ and discriminator $D_{\phi}$; Newly-generated storyline sequences by $\pi^{G_{\theta}}$ on unseen test dataset.}
    \tcp{Pre-processing}
    Derive embeddings $V_{e}$ on $\D$ and map entity $E$ in $V_{e}$\;
    Compute representations $V_{i}$ of based on $\D$ using multimodal learning, and map each of $I_{1:T}$ to $V_{i}$\;
    Generate the sequence ${\SL}$ including $V_{e}$, $V_{i}$ and $V_{i} - V_{e}$\;
    \tcp{Pre-training}
    Train $\pi^{G_{\theta}}$ using LSTM on $\SL$\;
    Generate negative samples using $\pi^{G_{\theta}}$ for training $D_{\phi}$\;
    Feed $D_{\phi}$ with negative samples and real data\;
    Train $D_{\phi}$ via minimizing the cross entropy\;
    \tcp{GAN training}
    \Repeat{GAN converges}{
        \For{$\pi^{G_{\theta}}$ training}
        {
            Sample a sequence $\SL_{1:T}^{\pi^{G_{\theta}}}=(s_{1},...,s_{T}) \sim \pi^{G_{\theta}}$\;
            \For{t in 1:T}{
                Derive $Q(s_{s};\SL_{1:t-1})$ using Eq. (\ref{mc})
            }
            Update generator: $\theta^{i} \leftarrow \theta+\alpha\nabla _{\theta }J\left( \theta \right)$
            
        }
        \For{$D_{\phi}$ training}
        {
            Generate negative examples by sampling $\SL_{1:T}^{\pi^{G_{\theta}}}=(s_{1},...,s_{T}) \sim \pi^{G_{\theta}}$\;
            Use given positive examples $\SL$\;
            Train discriminator $D_{\phi}$ by Eq. (\ref{gan})
        }
    }    
    Apply $\pi^{G_{\theta}}$ on unseen dataset $Data'$ starting from randomly selected entity $E' \in Data'$
\end{algorithm}

\subsection{Multimodal Learning}\label{ML}
The task contains three types of modalities which are text, image, and their coherence. Entity words are encoded using word embeddings algorithm such as Word2Vec. Whereas our model expresses images in a semantic sentential space instead of word space since images contain massive information.

A SVD-based semantic embedding model is employed to derive vectors for images conditioning on contextual words. Denote the image vector $\I \in M_{k}(\mathbb{R})$ from a multimodal vector space and is associated with sentential description $\mathbf \SEN = \{w_{1},...,w_{N}\}$ where $w_{i}$ indicates words. To represent the vectors $\I$ is to condition on the embedding vector of the description $\SEN$. We aim to model the conditional distribution $Pr(w_{n}|w_{1:n-1,}, \I)$ of the following word $w_{n}$ given context from the contextual words and the vector $\I$. Vocabulary embedding matrix with associated image vectors are represented using a tensor $\Psi= \I_{k} \otimes \SEN_{v\times k} \in M_{v\times k\times k}(\mathbb{R}) $ where $\otimes$ mean tensor product in which any word in description is associated with related images. Given $\I$, the model predicts word representation of the following words as a function of $\I$ and contextual words using tensor decomposition: 
\begin{equation}
    \Psi(\I) = U_{jv}^{T} \cdot \Sigma(\sigma_{jk},\I) \cdot V_{jk}
\end{equation}
 where the dimensionality of $j$ is tunable. $\Sigma$ denote a function which retain all the arguments on the diagonal after multiplication. This idea share resemblance with SVD in which the middle matrix is a diagonal matrix.  The context is then represented as an intermediate variable which subject to the expectation of weights distribution among word pairs : $C=\E_{v\times v}\left[U_{jk}^{T} V_{fv}({w_{i}})\right]$. Combining with $I$, we get another intermediate factor which encode context and image using Hadamard product $\psi=(UC) \cdot (V\I)$. Using with softmax, the conditional probability $Pr$ can be calculated. The true following words are used as true label for backpropagation method such as SGD, this neural networks iteratively update w.r.t. the parameters. The training parameters include $U_{jv},\sigma_{jk}, V_{jk}$, and weights distribution $\E_{vv}$. The this model encodes image in a vector space $V_{i}$. Finally, the third modality is the sequential difference between word embeddings and image vectors.

\section{Benchmark and Experiment Setting}\label{exp}
The proposed method is evaluated on newly-proposed storytelling dataset\footnote{\scriptsize{https://gist.github.com/aquastar/03dadfd751f5862ea0b44bb66996b490}}.
To guide the model to discover desirable stories, manually labeled storylines are compiled for GAN training. Generator obtained in one event dataset was tested on another event corpus. This experiment shows if the generator is capable of deriving transferable storyline. Please note that different event datasets share no entities.

\subsection{Benchmark Description and Metrics}
\textbf{Training set} contain events from two major categories: \textit{Homicide} and \textit{Protest}. \textit{Homicide} contains
1310 storylines, while \textit{Protest} chooses 934 storylines. Two short examples are shown in Figure \ref{MIS}. They are from the most famous historical events, such as \textit{9/11 Attack} , \textit{Orlando Nightclub Attack}, \textit{Occupy Wall Street}, and \textit{Protest led by Martin Luther King}. Each event contains several hundred documents. All the articles are taken from Google news and Wikipedia. Generators will be trained in the two categories separately. The \textbf{Test set} includes two event \textit{2014 Iguala mass kidnapping}\footnote{https://en.wikipedia.org/wiki/2014\_Iguala\_mass\_kidnapping} and \text{Malaysia Airlines Flight 370}\footnote{https://en.wikipedia.org/wiki/Malaysia\_Airlines\_Flight\_370}, since the two test events involve both \textit{homicide} and \textit{protest} sub-events. For data augmentation purpose, slicing window is employed to divide raw data into minimal sequence. 

\textbf{Metrics:} First the proposed method with several baselines are tested under convergence performance. Secondly, our evaluation conducts user studies via Amazon Mechanical Turk(AMT) cloud sourcing service, since the ultimate goal is to validate imitation behaviors.

\subsection{Training Setting}
Words are expressed using Word2Vec in each independent event corpus, while images are initially represented in VGG19 and then transferred to sentential space using multimodal learning. To normalize their shape, word vectors along each storyline are reduced by the vector of the first word in its storyline. Likewise, image vectors are reduced by the first image in each storyline. The same operation is also conducted on the multimodal sequence. For GAN model, the generator is implemented using LSTM model which accept continuous values. TextCNN\cite{zhang2015text} is used for discriminator since such CNN is effective for both text and image. Baselines include Random(\textbf{Ran.}), Scheduled sampling(\textbf{SS})\cite{bengio2015scheduled}, policy gradient(\textbf{PG}), and \textbf{LSTM}.

\subsection{Initial Analysis}
We evaluate our result with several established alternatives: random, scheduled sampling\cite{bengio2015scheduled} and policy gradient with similarity. Unfortunately, the baselines do not share the same metric or objective function. Instead, they were compared regarding accumulative normalized similarity on the training set:
\begin{equation*}
    sum_{sim} = \sum _{\A\in Data}\sum _{\B\in model}\dfrac {\A\B}{\left\| \A\right\| _{F}\cdot \left\| \B\right\| _{F}}
\end{equation*}
where, sample $\A$ denotes sequence matrix from the user-provided data, and $\B$ is the output sequence matrix from models. The result is shown in Table (\ref{sim_base}).

\begin{table}
\caption{Similarity performance (T./I. denotes Text/Image respectively, T.I. means the combination of T. and I.)}
\begin{center}
\label{sim_base}
 \begin{tabular}{|c |c |c  |c |}
 \hline
   Ran. & SS & PG & LSTM\\
 \hline
   -3860.02   & 33714.67 & 34009.27 & 338876.34  \\ 
 \hline
   T. & I. & T.I. & MIL-GAN\\
 \hline
   34263.59 & 12483.20   & 36143.34 & \textbf{36697.33}  \\ 
 \hline
\end{tabular}
\end{center}
\end{table}

Using the multimodal GAN, we improve the text-only model. One interesting point is that the image-only modality does not performance very well, but with other modalities, the performance increases. This makes sense because an image may contain too much information which is likely to lead to confusion. For instance, an image of original World Trade Center being attached could be either a \textit{victim} object(the building) or an \textit{attacker} (the attacking flight). With textual label such as victim, the image would be confined in a smaller and accurate semantic space.

The balance parameters $\lambda_{i=1,2,3}$ are all initialized to 1. After fine tuning, good performance often appear if more weights were assigned on text part. One good set example is [0.6, 0.3, 0.1] for [$V_{e}$, $V_{i}$ , $V_{i} - V_{e}$] separately.

However, result of the training set can only show that the potential ability to achieve success, but it is not the ultimate evaluation. In next section, we conducted user study via Amazon Mechanical Turk(AMT) Service to directly assess users' satisfaction.

\subsection{Evaluation via user study}\label{case}
Evaluating storytelling is a difficult task, due to the fact that there is no established golden standard, and even ground truth is hard to elicit. The ultimate goal of imitation storytelling is to help users discover customized storylines given users' examples. Therefore, a third-party user study via AMT is conducted, since it allows us to obtain accurate statistical information with a large sample size of users. It aims to test whether the generated storyline matches users' interests. We evaluate our result with the baselines mentioned in the last subsection. 

In the study, workers were asked which storyline fit the given storyline best. Due to our dynamic reward per assignment in AMT, every minimal unit of cost is treated as one evaluation, so 1000 rating were collected. For each test event, several starting entities were randomly chosen. Storylines were generated from these starting entities using both our proposed method and baselines. For each unit task in AMT, workers were asked to choose the best one among those generated candidates gave event background knowledge such as Wikipedia or key news articles:
{\small
\begin{itemize}
    \item \textbf{1st Step}: AMT will be given E0, which denotes an event background and user-provided storyline S0 which is a user-define storyline. S0 contain a topic which is based on E0.

    \item \textbf{2nd Step}: After AMT workers study the relationship between the storyline and event, they will be given another event E1 and a few machine generated storyline S1/S2/S3/S4. They are also related.
    \item \textbf{3rd Step}: Based on their understanding of the relationship S0-E0, choose the best one that would have similar relationship: E1-S[?] among (S1,S2,S3,S4) 
\end{itemize}
}
The final result as shown in Table \ref{amt} demonstrates that our proposed MIL-GAN was significantly preferred by users. T. slightly improve the preference beyond Schedule Sampling. Policy Gradient fails to capture the users' behavior.

\begin{table}
\caption{Preference statistics from AMT}
\begin{center}
\label{amt}
 \begin{tabular}{|c c |c c|}
 \hline
    PG & SS & T.& MIL-GAN\\
 \hline
    17.3\% & 23.2\% & \textbf{28.1\%} & \textbf{31.4\%} \\ 
 \hline
\end{tabular}
\end{center}
\end{table}

\vspace{-10pt}
\subsection{case analysis}
In this subsection, several typical examples are presented, and detailed analysis will be elaborated.
\subsubsection{case 1: Mexico Murderer}
{\small
\begin{itemize}
    \item \textbf{SS}: Abarca$\rightarrow$Flipe$\rightarrow$Flipe$\rightarrow$Guerrero
    \item \textbf{PG}: Abarca$\rightarrow$Pineda$\rightarrow$student$\rightarrow$student
    \item \textbf{T.}: Abarca$\rightarrow$gang$\rightarrow$student$\rightarrow$Guerrero
    \item \textbf{MIL-GAN}: Abarca$\rightarrow$Flipe$\rightarrow$Ayotzinapa$\rightarrow$Guerrero
\end{itemize}
\subsubsection{case 1: Mexico Drug Cartel}
\begin{itemize}
    \item \textbf{SS}: Cartel$\rightarrow$Flipe$\rightarrow$Flipe$\rightarrow$Guerrero
    \item \textbf{PG}: Cartel$\rightarrow$gang$\rightarrow$gang$\rightarrow$Sinaloa
    \item \textbf{T.}: Cartel$\rightarrow$Drug$\rightarrow$Pineda$\rightarrow$gang
    \item \textbf{MIL-GAN}: Cartel$\rightarrow$Pineda$\rightarrow$Abarca$\rightarrow$Flipe
\end{itemize}
\subsubsection{case 2: MH370}
\begin{itemize}
    \item \textbf{SS}: Zaharie$\rightarrow$wife$\rightarrow$wife$\rightarrow$cockpit
    \item \textbf{PG}: Zaharie$\rightarrow$Flight$\rightarrow$Fariq$\rightarrow$training
    \item \textbf{T.}: Zaharie$\rightarrow$Flight$\rightarrow$Aisa$\rightarrow$Search
    \item \textbf{MIL-GAN}: Zaharie$\rightarrow$MH370$\rightarrow$India$\rightarrow$debris
\end{itemize}
}

We found that \textbf{(1)}The generated sequences from the baselines \textbf{SS} and \textbf{PG} often contains more vague words such as \textit{student} and \textit{gang}. These two baselines tend to overfit and often yield repeat nodes as in case 1 and 2. This suggests that the models without GAN are likely to overfit. \textbf{(2)} With GAN, the proposed models seems succeeded to avoid the overfit issue. Compared with \textbf{SS} and \textbf{PG}, \textbf{T.} and \textbf{MIL-GAN} tend to generate specific word rather than ambiguous ones. For example, in case 1, 2 and 3, \textbf{T.} and \textbf{MIL-GAN} generated specific words such as \textit{Aisa} and \textit{debris}. Compared with \textbf{T.}, \textbf{MIL-GAN} generates even more specific entity name such as \textit{Ayotzinapa}, \textit{Flipe} and \textit{India}. This implies that multimodal constraints does improve the unimodal performance.   

\section{Conclusion}\label{conclusion}
In this paper, we proposed a multimodal imitation learning approach for generating storyline on unseen events. To avoid the reward function designing, GAN based imitation learning is introduced to learn the latent policy given users' demonstrations. To bridge the information gap between text and image, our model effectively integrates generative adversarial nets and multimodal learning via deterministic policy gradient. The different modalities learn from each other and potentially resolve confusion from each single modality. In our experiments, we utilized user-provided demonstration to explicitly illustrate the advantage of the proposed method beyond baselines. Associating with multimodal perspective, our model succeeds to capture the latent patterns across different modalities, and therefore reveal more satisfying storylines towards users' interests.


\bibliographystyle{named}
\bibliography{ijcai17}
\end{document}